\documentclass{article}

\usepackage{arxiv}

\usepackage[utf8]{inputenc} 
\usepackage[T1]{fontenc}    
\usepackage{hyperref}       
\usepackage{url}            
\usepackage{booktabs}       
\usepackage{amsmath,amssymb} 
\usepackage{amsfonts}       
\usepackage{nicefrac}       
\usepackage{microtype}      
\usepackage{cleveref}       
\usepackage{lipsum}         
\usepackage{graphicx}
\usepackage[square]{natbib}
\usepackage{doi}
\usepackage{mathtools}
\usepackage{color}
\usepackage{bbm}
\usepackage{siunitx}
\usepackage{multirow}
\usepackage[acronym]{glossaries}

\newcommand{\ie}{\textit{i}.\textit{e}., }
\newcommand{\eg}{\textit{e}.\textit{g}., }
\DeclareMathOperator*{\argmax}{arg\,max}

\newacronym[plural=WSIs,firstplural=Whole-Slide Images (WSIs)]{wsi}{WSI}{Whole-Slide Image}
\newacronym{he}{H\&E}{Hematoxylin \& Eosin}
\newacronym{cp}{CompPath}{Computational Pathology}
\newacronym{dp}{DigPath}{Digital Pathology}
\newacronym{dl}{DL}{Deep Learning}
\newacronym[plural=RNNs,firstplural=Recurrent Neural Networks (RNNs)]{rnn}{RNN}{Recurrent Neural Network}
\newacronym[plural=GNNs,firstplural=Graph Neural Networks (GNNs)]{gnn}{GNN}{Graph Neural Network}
\newacronym[plural=MLPs,firstplural=Multi-Layer Perceptrons (MLPs)]{mlp}{MLP}{Multi-layer Perceptron}
\newacronym[plural=CNNs,firstplural=Convolutional Neural Networks (CNNs)]{cnn}{CNN}{Convolutional Neural Network}
\newacronym[plural=GAs,firstplural=Gated Attention modules (GAs)]{ga}{GA}{Gated Attention}
\newacronym[plural=DGAs,firstplural=Dual Gated Attention modules (GAs)]{dga}{DGA}{Dual Gated Attention}
\newacronym{mil}{MIL}{Multiple Instance Learning}
\newacronym{flop}{FLOP}{Floating-Point Operation}
\newacronym[plural=RoIs,firstplural=Regions-of-Interest (ROIs)]{roi}{RoI}{Region-of-Interest}
\newacronym[plural=TRoIs,firstplural=Tumor Regions-of-Interest (TRoIs)]{troi}{TRoI}{Tumor Region-of-Interest}

\makeglossaries

\newcommand{\randomps}{\textsc{Random}}
\newcommand{\nondiffkps}{\textsc{NonDiff-TopK}}
\newcommand{\difftopkps}{\textsc{Diff-TopK}}

\newcommand{\clam}{\textsc{CLAM}}
\newcommand{\sparseconvmil}{\textsc{SparseConvMIL}}
\newcommand{\transmil}{\textsc{TransMIL}}
\newcommand{\abmil}{\textsc{ABMIL}}
\newcommand{\maxmil}{\textsc{MaxMIL}}
\newcommand{\meanmil}{\textsc{MeanMIL}}
\newcommand{\ours}{\textsc{ZoomMIL}}

\title{Differentiable Zooming for Multiple Instance Learning on Whole-Slide Images}

\date{}

\author{
    Kevin Thandiackal$^{\mathbf{1,2}}$\thanks{The authors contributed equally to this work.} ,
    Boqi Chen$^{\mathbf{1,2}*}$,
    Pushpak Pati$^\mathbf{1}$,
    Guillaume Jaume$^\mathbf{3}$\thanks{Work done while at IBM Research Europe}\\
	\textbf{Drew F. K. Williamson$^\mathbf{3}$,
    Maria Gabrani$^\mathbf{1}$,
    Orcun Goksel$^\mathbf{2,4}$}
    \\[1.5ex]
    $^1$IBM Research Europe, Zurich, Switzerland\\
    $^2$Computer-assisted Applications in Medicine, ETH Zurich, Zurich, Switzerland\\
    $^3$Brigham and Women’s Hospital, Harvard Medical School, Boston, USA\\
    $^4$Department of Information Technology, Uppsala University, Uppsala, Sweden\\
}

\renewcommand{\shorttitle}{Differentiable Zooming for Multiple Instance Learning on Whole-Slide Images}

\begin{document}
\maketitle

\begin{abstract}
    Multiple Instance Learning (MIL) methods have become increasingly popular for classifying giga-pixel sized Whole-Slide Images (WSIs) in digital pathology.
    Most MIL methods operate at a \emph{single} WSI magnification, by processing \emph{all} the tissue patches.
    Such a formulation induces high computational requirements, and constrains the contextualization of the WSI-level representation to a single scale. A few MIL methods extend to multiple scales, but are computationally more demanding. In this paper, inspired by the pathological diagnostic process, we propose \textsc{ZoomMIL}, a method that \emph{learns} to perform multi-level zooming in an end-to-end manner. 
    \textsc{ZoomMIL} builds WSI representations by aggregating tissue-context information from multiple magnifications.
    The proposed method outperforms the state-of-the-art MIL methods in WSI classification on two large datasets, while significantly reducing the computational demands with regard to Floating-Point Operations (FLOPs) and processing time by up to 40$\times$.
\end{abstract}


\section{Introduction}
\begin{figure}[!t]
    \centering
    \includegraphics[width=\textwidth]{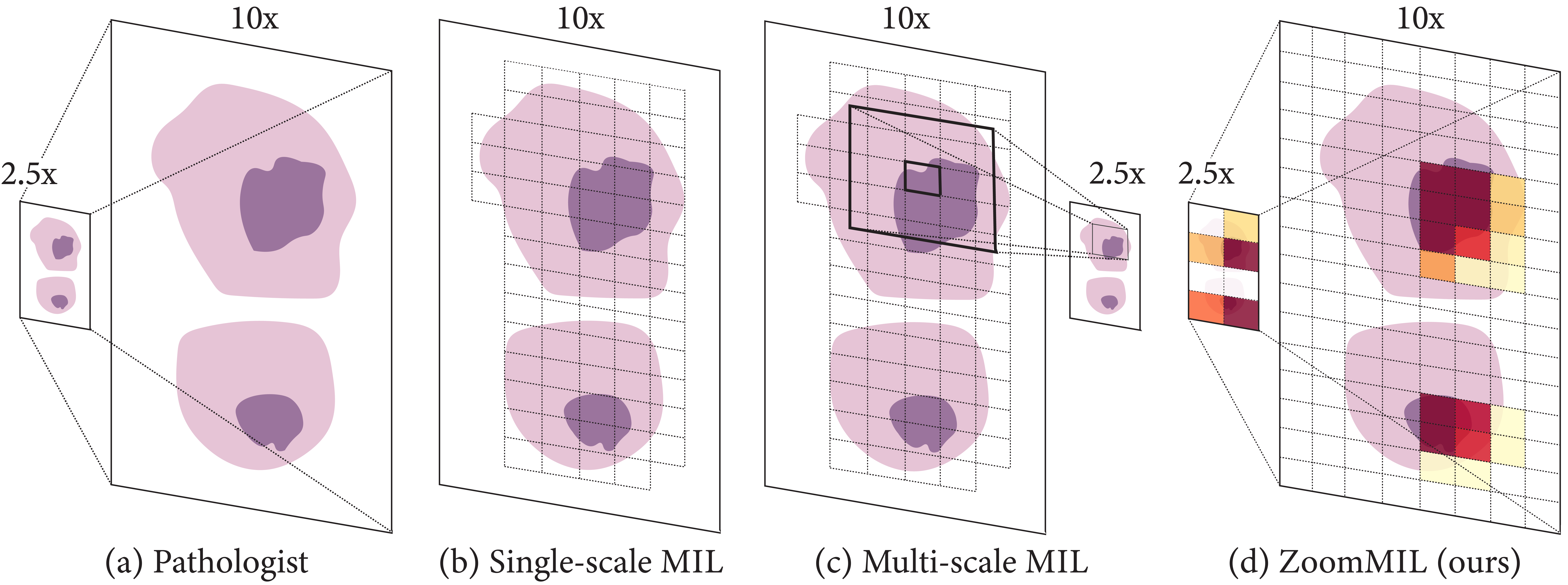}
    \caption{
    Comparison of different methods for the diagnosis of \glspl*{wsi}
    }
    \label{fig:multiscale_analysis}
\end{figure}

Histopathological diagnosis requires a pathologist to examine tissue to characterize the phenotypic, morphologic, and topologic distribution of the tissue constituents.
With advancements in slide-scanning technologies, tissue specimens can now be digitized into \glspl*{wsi} with impressive resolution, enabling the pathological assessment to be conducted on a computer screen rather than under a microscope. As does a glass slide, a \gls*{wsi} contains rich tissue information, and can be up to 100\,000$\times$100\,000 pixels in size at 40$\times$ magnification (0.25$\mu$m/pixel).
Due to image size and complexity as well as the multi-scale nature of biological systems, a pathologist generally examines the slide in a hierarchical manner, beginning with the detection of informative regions at a \emph{low} magnification, followed by a detailed evaluation of selected areas at a \emph{high} magnification, as exemplified in Figure~\ref{fig:multiscale_analysis}(a).
However, depending on the tissue sample, such manual examination of a gigapixel-sized \gls*{wsi} can be cumbersome, time-consuming, and prone to inter- and intra-observer variability~\citep{gomes14,elmore15}.

To alleviate the aforementioned challenges, \gls*{dl} based diagnosis tools are being developed in digital pathology. However, these tools encounter additional challenges pertaining to the large size of the \glspl*{wsi}, and the difficulty of acquiring fine-grained, pixel-level annotations.
To address these, dedicated \gls*{dl} methods have been proposed, in particular based on \gls*{mil}.
In \gls*{mil}, a \gls*{wsi} is first decomposed into a bag of patches, which are individually encoded by a backbone \gls*{cnn}.
Then, a pooling operation combines the resulting patch embeddings to produce a slide-level representation that is finally mapped to the bag/image label.
Although these approaches have achieved remarkable performance on a variety of pathology tasks, such as 
tumor classification~\citep{campanella19,lu20b,tellez20,shaban20}, 
tumor segmentation~\citep{jia2017,liang2018weakly}, 
and survival prediction~\citep{yao21}, they pose the following drawbacks.

First, the performance of an \gls*{mil} method relies on a carefully tuned context-resolution trade-off~\citep{pati21,bejnordi17,sirinukunwattana18}, \ie an optimal resolution of operation while including adequate context in a patch. As the dimensions of diagnostically relevant tissue constituents vary significantly in histopathology, patches of different sizes at different magnifications render different context information about the tissue microenvironment. Thus, identifying an optimal resolution and patch size involves several tailored and tedious steps.
Typical \gls*{mil} methods work with patches at a \emph{single} magnification, as shown in Figure~\ref{fig:multiscale_analysis}(b), and disregard the spatial distribution of the patches, leading to the above problem.
Though some \gls*{mil} methods address this trade-off via visual self-attention~\citep{myronenko21,shao21}, they are constrained by the expensive computation of attention scores on a large number of patches in a \gls*{wsi}.
Differently, \citep{lerousseau2021sparseconvmil} addresses the issue via random patch sampling and sparse-convolutions, which consequently prevents deterministic inference.
In the literature, other methods~\citep{sirinukunwattana18,hashimoto20,ho21} are proposed based on the concept of concentric-patches across multiple magnifications, as shown in Figure~\ref{fig:multiscale_analysis}(c), to acquire richer context per patch. However, these methods are computationally more expensive as they require to encode all the patches at a high magnification and all the corresponding patches across other lower magnifications.

Second, \gls*{mil} approaches process \emph{all} the tissue patches at a high magnification. Thus, a large number of uninformative patches are processed, increasing the computational cost, inference time, and memory requirements.
For instance, inferring on a \gls*{wsi} of $50\,000 \times 50\,000$ pixels by $\clam$~\citep{lu20b}, an \gls*{mil} method, requires $\approx$150 Tera \glspl*{flop}, which is 37\,500$\times$ the processing of an ImageNet~\citep{deng2009imagenet} sample by  ResNet34~\citep{he2016deep}.
Further, the high memory footprint of these methods inhibits their scalability to large histopathology images, \eg prostatectomy slides which can be of 300\,000$\times$400\,000 pixels at 40$\times$ magnification.
Such requirements of intensive computational resources, can in turn hinder the clinical deployment of these approaches. Their adoption becomes even prohibitive when the computational resources are scarce owing to a limited access to GPUs or cloud services.
In view of the above challenges with the existing \gls*{mil} methods, a multi-scale context-aware \gls*{mil} method with a high computational efficiency is desired.

In this paper we propose $\ours$, a novel method inspired by the hierarchical diagnostic process of pathologists, to address the above drawbacks with \gls*{mil}.
We first identify \glspl*{roi} at a low magnification, and then zoom in on them at a high magnification for fine-grained analysis, as shown in Figure~\ref{fig:multiscale_analysis}(d). The \gls*{roi} selection is performed through a gated-attention and a differentiable top-K ($\difftopkps$) module, which allows to learn where to zoom in, in an end-to-end manner, while moderating the computational requirements at high magnifications.
This process can be repeated across an arbitrary number of magnifications, \eg 5$\times$$\rightarrow$10$\times$$\rightarrow$ 20$\times$, as per the application at hand.
Finally, we aggregate the information acquired across multiple scales to obtain a context-aware \gls*{wsi} representation to perform downstream pathology tasks, as presented in Figure~\ref{fig:Overview}. 
In summary, our contributions are:
\begin{enumerate}
    \item A novel multi-scale context-aware \gls*{mil} method that learns to perform multi-level zooming in an end-to-end manner for \gls*{wsi} classification.
    
    \item A computationally much more efficient method compared to the state of the art in \gls*{mil}, \eg 40$\times$ faster inference on a \gls*{wsi} of size 26\,009$\times$18\,234 pixels at 10$\times$ magnification, while achieving better (2/3 datasets) or comparable (1/3 datasets) \gls*{wsi} classification performance.
    
    \item Comprehensive benchmarking of the method with regard to \gls*{wsi} classification performance and computational requirements (on GPU and CPU) on multiple datasets across multiple organs and pathology tasks, \ie tumor subtyping, grading, and metastasis detection.
\end{enumerate}

\section{Related Work}

\subsection{Multiple Instance Learning in Histopathology}

\gls*{mil} in histopathology was introduced in~\citep{ilse18} to classify breast and colon \glspl*{roi}. The experiments established the superiority of \emph{attention}-based pooling over vanilla \emph{max} and \emph{mean} pooling operations.
Concurrently, \citep{campanella19} scaled \gls*{mil} to \gls*{wsi}-level for grading prostate needle biopsies. This work proposed the use of end-to-end training with \gls*{rnn}-based pooling. 
Later, several works consolidated attention-based \gls*{mil} by applying it across several organs and pathology tasks, \eg \citep{yao21,lu20b,lu21}.
Recently, transformer-based \gls*{mil} methods~\citep{myronenko21,shao21} were proposed to take into account the inter-patch dependencies, with the downside of having to compute a quadratic number of interactions, which introduces memory constraints.
Additionally, all the aforementioned \gls*{mil} methods are limited to operate on \emph{all} the patches in a \gls*{wsi} at a single magnification. 
In view of the benefits of multi-scale information in histopathology image analysis~\citep{bejnordi15,gao16,tokunaga19,li20,ho21,pati21}, a few recent methods~\citep{hashimoto20,li21} were proposed to extend the \gls*{mil} methods to combine information across multiple magnifications.
However, similar to the single-scale methods, these multi-scale versions also require to process \emph{all} the patches in a \gls*{wsi}, which is computationally more expensive.
In contrast, our proposed $\ours$ \emph{learns} to identify informative regions at low magnification, and subsequently zooms into these regions at high magnification for efficient and comprehensive analysis. 
Differently, several other approaches aim to learn the inter-instance relations in histopathology via \glspl*{gnn}~\citep{aygunes20,pati21,zhao20,raju20,li18,adnan20,anklin21} or \glspl*{cnn}~\citep{tellez20,shaban20,lerousseau2021sparseconvmil}.

\subsection{Instance Selection Strategies in Histopathology}

Most \gls*{mil} methods encode all the patches in a \gls*{wsi} irrespective of their functional types.
This definition compels \gls*{mil} to be computationally expensive for large \glspl*{wsi} in histopathology. 
To reduce the computational memory requirement, \citep{lerousseau2021sparseconvmil} proposed to randomly sample a subset of instances, with the consequence of potentially missing vital information, especially when the informative set is small, \eg metastasis detection.
Differently, a few reinforcement learning-based methods~\citep{dong18,qaiser19} were proposed to this end. \citep{qaiser19} proposed to sequentially identify some of the diagnostically relevant \glspl*{roi} in a \gls*{wsi} by following a parameterized policy. However, the method leverages a very coarse context information for the \gls*{roi} identification, and is limited to utilize only single-scale information for the diagnosis.
Additionally, the reinforcement learning method in \citep{dong18} and the recurrent visual attention-based model in \citep{bentaieb18} aim to select patches which mimics pathological diagnosis. However, these methods require pixel-level annotations for learning discriminative regions, which is expensive to acquire on large \glspl*{wsi}. In contrast to the above methods, $\ours$ requires only \gls*{wsi}-level supervision.
Our method is flexible to attend to several magnifications, while efficiently classifying \glspl*{wsi} with high performance.

The attention-score-based iterative sampling strategy proposed in \citep{kong21,katharopoulos19} closely relates to our work. For the final classification, the selected patch embeddings are simply concatenated, analogous to average pooling. 
Instead, $\ours$ incorporates a dual gated-attention module between two consecutive magnifications to simultaneously learn to select the relevant instances to be zoomed into, and learn an improved \gls*{wsi}-level representation for the lower magnification.

The patch selection module employed in our work is inspired by the perturbed optimizer-based~\citep{berthet2020learning} differentiable Top-K algorithm proposed in~\citep{cordonnier21}.
$\ours$ advances upon~\citep{cordonnier21} by extending to several magnifications, \ie multi-level zooming, and scaling the applications to giga-pixel sized \glspl*{wsi}.

\section{MIL with Differentiable Zooming}

In this section, we present our differentiable zooming approach, that first identifies informative patches at a low magnification and afterwards zooms into them for fine-grained analysis.
In Sec.~\ref{sec:att_MIL}, we introduce the gated-attention mechanism that determines the informative patches at a magnification.
In Sec.~\ref{sec:att_high_mag}, we describe how to enable the attention-based patch selection to be differentiable while employing multiple magnifications.
Finally, we present in Sec.~\ref{sec:dual_ga} our overall architecture, in particular our proposed Dual Gated Attention and multi-scale information aggregation.

\subsection{Attention-based MIL} \label{sec:att_MIL}

In \gls*{mil}, an input $X$ is considered as a bag of multiple instances, denoted as $X = \{\mathbf{x}_1, ..., \mathbf{x}_N\}$.
Given a classification task with $C$ labels, there exists an \emph{unknown} label $\mathbf{y}_i \in C$ for each instance and a \emph{known} label $\mathbf{y} \in C$ for the bag.
In our context, the input is a \gls*{wsi} and the instances denote the set of extracted patches.
We follow the embedding-based \gls*{mil} approaches~\citep{ilse18,lu20b,shao21}, where a patch-level feature extractor $h$ maps each patch $\mathbf{x}_i$ to a feature vector $\mathbf{h}_i = h(\mathbf{x}_i) \in \mathbb{R}^D$.
Afterwards, a pooling operator $g(\cdot)$ aggregates the feature vectors $\mathbf{h}_{i=1:N}$ to produce a single \gls*{wsi}-level feature representation.
Finally, a classifier $f(\cdot)$ uses the \gls*{wsi} representation to predict the \gls*{wsi}-level label $\hat{\mathbf{y}} \in C$. The end-to-end process can be summarized as:
\begin{equation} \label{eq:mil_funcs}
    \hat{\mathbf{y}} = f \bigg( g \Big( \{ h(\mathbf{x}_1), \dots , h(\mathbf{x}_N) \} \Big) \bigg) \;.
\end{equation}

To aggregate the patch features, we employ an attention-pooling operation, specifically, the \gls*{ga} mechanism introduced in~\citep{ilse18}.
Let $\mathbf{H} = [\mathbf{h}_1, \dots, \mathbf{h}_N]^\top \in \mathbb{R}^{N \times D}$ be the patch-level feature matrix, then the \gls*{wsi}-level representation $\mathbf{g}$ is computed as:
\begin{equation}
    \mathbf{g} = \sum_{i=1}^{N} a_i \mathbf{h}_i, \qquad a_i = \frac{\exp\{\mathbf{w}^\top (\tanh(\mathbf{V}\mathbf{h}_i) \odot \eta(\mathbf{U} \mathbf{h}_i))\}}{\sum_{j=1}^N \exp\{ \mathbf{w}^\top (\tanh(\mathbf{V}\mathbf{h}_j) \odot \eta(\mathbf{U} \mathbf{h}_j)) \}} \;,
\end{equation}
where $\mathbf{w}$$\in$$\mathbb{R}^{L \times 1}$, $\mathbf{V}$$\in$$\mathbb{R}^{L \times D}$, 
$\mathbf{U}$$\in$$\mathbb{R}^{L \times D}$ are learnable parameters with hidden dimension $L$, $\odot$ is element-wise multiplication, and $\eta(\cdot)$ is the sigmoid function.
While previous attention-based \gls*{mil} methods~\citep{ilse18,lu20b} are designed to operate at only a \emph{single} magnification, we propose an efficient and flexible \gls*{mil} framework that can be extended to an arbitrary number of magnifications while being fully differentiable.

\begin{figure}[!t]
    \centering
    \includegraphics[width=\textwidth]{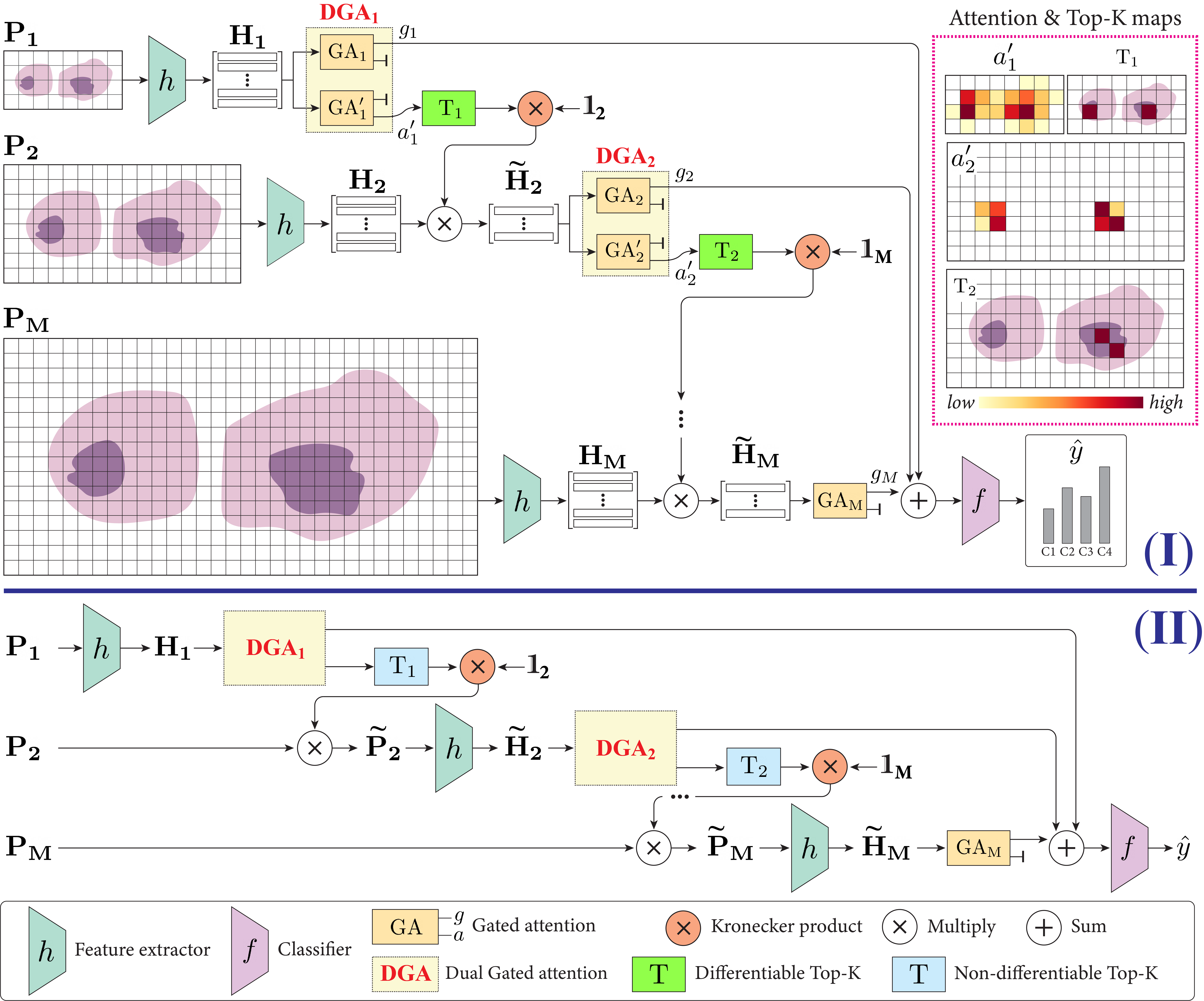}
    \caption{Overview of the proposed $\ours$. \textbf{(I)} and \textbf{(II)} present the distinct training and inference modes, generically exemplified for $M$ magnifications
    }
    \label{fig:Overview}
\end{figure}

\subsection{Attending to Multiple Magnifications}\label{sec:att_high_mag} 

Let us assume the \gls*{wsi} is accessible at multiple magnifications, indexed by $m \in \{1, 2, \dots, M\}$, where $M$ depicts the highest magnification.
Consistent with the pyramidal format of \glspl*{wsi}, we assume the magnification at $m+1$ is twice the magnification at $m$.
To efficiently extend \gls*{mil} to multiple magnifications, we propose to hierarchically identify informative patches from low-to-high magnifications, and aggregate their features to get the \gls*{wsi} representation.
To identify the informative patches at $m$, we first compute $\mathbf{a}_m \in~\mathbb{R}^N$, that includes an attention score for each patch.
Then, the top $K$ patches with the highest attention scores are selected for further processing at a higher magnification. The corresponding selected patch feature matrix can be denoted by, 
\begin{equation}
    \widetilde{\mathbf{H}}_{m} = \mathbf{T}^\top_{m} \mathbf{H}_{m} \;,
    \label{eq:patch_selection}
\end{equation}
where $\mathbf{T}_{m} \in \{0,1\}^{N \times K}$ is an indicator matrix, and $\mathbf{H}_{m} \in \mathbb{R}^{N \times D}$ is the patch feature matrix at $m$.

Instead of a handcrafted approach, we propose to drive the patch selection at $m$ directly by the prediction output of $f(\cdot)$. 
This can be trivially achieved via a backpropagation path from the output of $f(\cdot)$ to the attention module at $m$, without introducing any additional loss or associated hyperparameters.
However, this naive formulation is non-differentiable as it involves a Top-K operation.
To address this problem, we build on the perturbed maximum method~\citep{berthet2020learning} to make the Top-K selection differentiable, inspired by~\citep{cordonnier21}, and apply it to the attention weights $\mathbf{a}_{m}$  at magnification $m$. 
Specifically, $\mathbf{a}_{m}$ is first perturbed by adding uniform Gaussian noise $\mathbf{Z} \in \mathbb{R}^N$.
Then, a linear program is solved for each of these perturbed attention weights, and their results are averaged. Thus, the forward pass of the differentiable Top-K module can be written as:
\begin{equation}
    \mathbf{T} = \mathop{\mathbb{E}}_{\mathbf{Z}\sim \mathcal{N}(0, \mathbbm{1})} \Big[ \argmax_{\hat{\mathbf{T}}} \langle \hat{\mathbf{T}}, \big( \mathbf{a}_m + \sigma \mathbf{Z} \big) \mathbf{1}^\top \rangle \Big] \;,
    \label{eq:perturbed_topk}
\end{equation}
where $\mathbf{1}^\top = [1 \cdots 1] \in \mathbb{R}^{1 \times K}$ and consequently $(\mathbf{a}_m + \sigma \mathbf{Z})\mathbf{1}^\top \in \mathbb{R}^{N \times K}$ denotes the perturbed attention weights repeated $K$ times, and $\langle \cdot \rangle$ is a scalar product preceded by a vectorization of the matrices.
The corresponding Jacobian is defined as:
\begin{equation}
    J_{\mathbf{a}_m}\mathbf{T} = \mathop{\mathbb{E}}_{\mathbf{Z}\sim \mathcal{N}(0, \mathbbm{1})} \Big[ \argmax_{\hat{\mathbf{T}}} \langle \hat{\mathbf{T}}, \big( \mathbf{a}_m + \sigma \mathbf{Z} \big) \mathbf{1}^\top \rangle \mathbf{Z}^\top / \sigma \Big] \;.
    \label{eq:bwd_topk}
\end{equation}
More details on the derivation are provided in the supplemental material.
The differentiable Top-K operator enables to \emph{learn} the parameters of the attention module that weighs the patches at the specific magnification.
Notably, we maintain a constant patch size across all magnifications, unlike \citep{cordonnier21}, which proportionately scales the patch sizes across magnifications.
This proportionates the number of patches to the magnifications of operation.
It also allows different fields-of-view of the tissue microenvironment across different magnifications, and enables to capture a variety of context information. This is a crucial requirement for analyzing histopathology slides as they contain diagnostically relevant constituents of various sizes.
To achieve the objective of zooming, we propose to expand the indicator matrix $\mathbf{T}_m$ in order to select from the patch features $\mathbf{H}_{m'}~\in \mathbb{R}^{N \cdot 4^{(m'-1)} \times D}$, where $m' > m$.
Specifically, we compute the Kronecker product between $\mathbf{T}_{m}$ and the identity matrix $\mathbbm{1}_{m'} = \mathrm{diag}(1, \cdots, 1) \in \mathbb{R}^{4^{(m'-1)} \times 4^{(m'-1)}}$ to yield the expanded indicator matrix $\mathbf{T}_{m'} \in \{0, 1\}^{N \cdot 4^{(m'-1)} \times K \cdot 4^{(m'-1)}}$.
Analogously to Eq.~\eqref{eq:patch_selection}, the patch selection at higher magnification $m'$ using the attention weights from a lower magnification $m$ can be performed by using,
\begin{equation}
    \widetilde{\mathbf{H}}_{m'} = \big( \mathbf{T}_{m} \otimes \mathbbm{1}_{m'} \big)^\top \mathbf{H}_{m'},
    \label{eq:difftopk_matmul}
\end{equation}
where $\mathbf{H}_{m'}$ is the feature matrix at $m'$ and $\widetilde{\mathbf{H}}_{m'}$ is the selected feature matrix.

\subsection{Dual Gated Attention and Multi-Scale Aggregation} \label{sec:dual_ga}  
An overview of our method is presented in Figure~\ref{fig:Overview}.
The model has two distinct operational modes for training and inference, indicated as \textbf{(I)} and \textbf{(II)}.

\textbf{Training mode:} During training, the feature matrix $\mathbf{H}_{1}$ at $m$=1 passes through a \gls*{dga} block.
\gls*{dga} consists of two separate gated-attention modules $\textrm{GA}_{1}$ and $\textrm{GA}'_{1}$.
$\textrm{GA}_{1}$ is trained to obtain an optimal attention-pooled \gls*{wsi}-level representation $\bf{g}_1$ at low magnification.
$\textrm{GA}'_{1}$ aims to provide meaningful attention weights $\bf{a}'_1$ to facilitate the patch selection for the next higher magnification.
Alternatively, a single attention module could be employed to address both the tasks. 
However, this prevents the notion of optimal zooming, as the selected patches at the low magnification, to be zoomed-in, aim to optimize the classification performance of operating at the low magnification.
Employing separate attention modules decouples the optimization tasks, and in turn, enables to obtain \emph{complementary} information from both magnifications.
Subsequently, the differentiable Top-K selection module, $\textbf{T}_1$, is employed to learn to select the most informative patches. The following selected higher magnification patch feature matrix $\widetilde{\mathbf{H}}_{2}$ is obtained via Eq.~\eqref{eq:difftopk_matmul}.

The process of selecting patch features for every subsequent higher magnification is repeated until the highest magnification level $M$ is reached.
The selected patch features $\widetilde{\mathbf{H}}_{M}$ at $M$ go through a last gated-attention block $\textrm{GA}_{M}$ to produce the high magnification feature representation $\mathbf{g}_{M}$.
Finally, the attention-pooled feature representations from all magnifications, $\mathbf{g}_{1}, \mathbf{g}_{2}, \dots, \mathbf{g}_{M}$, are aggregated via sum-pooling to get a multi-scale, context-aware representation for the \gls*{wsi}. 
Inspired by residual learning~\citep{he2016deep}, sum-pooling is employed as the features across different  magnifications are closely related and the summation utilizes their complementarity.
The final classifier~$f(\cdot)$ maps the \gls*{wsi} representation to the \gls*{wsi} label $y \in C$, and produces the model prediction $\hat{\bf{y}}$.
The training phase can be regarded as extending Eq.~\eqref{eq:mil_funcs} with sum-pooling over multiple magnifications:
\begin{equation} \label{eq:sum_pooling}
    \hat{\mathbf{y}} = f \Big( \mathbf{g}_{1}(\mathbf{H}_{1}) + \mathbf{g}_{2}(\widetilde{\mathbf{H}}_{2}), \dots + \mathbf{g}_{M}(\widetilde{\mathbf{H}}_{M}) \Big) \;.
\end{equation}

\textbf{Inference mode:}
We introduce the differentiable Top-K operator in our model during training to learn to identify informative patches. However, this operator includes random perturbations to the attention weights, and consequently makes the forward pass of the model non-deterministic.
Therefore, we replace the differentiable Top-K with the conventional non-differentiable Top-K operator.
Additionally, non-differentiable Top-K is faster as no perturbations have to be computed.
As shown in Figure~\ref{fig:Overview}, another crucial difference to the training mode is that the patch selection directly operates on the \gls*{wsi} patches, $\mathbf{P}_{m'} \in \mathbb{R}^{N \cdot 4^{(m'-1)} \times p_h \times p_w \times p_c}$, instead of the pre-extracted patch features $\mathbf{H}_{m'}$.
This avoids the extraction of features for uninformative patches during inference, unlike other \gls*{mil} methods. It significantly reduces the computational requirements and speeds up model inference.

\section{Experiments}

\subsection{Datasets}
\label{datasets}
We benchmark our $\ours$ on three H\&E stained, public \gls*{wsi} datasets.

\textbf{CRC}~\citep{oliveira21} contains 1133 colorectal biopsy and polypectomy slides from \emph{non-neoplastic}, \emph{low-grade}, and \emph{high-grade} lesions, accounting for 26.5\%, 48.7\%, 24.8\% of the data. The slides were acquired at the IMP Diagnostics laboratory, Portugal, and were digitized by a Leica GT450 scanner at 40$\times$. We split the data into 70\%/10\%/20\% stratified sets for training, validation, and testing.

\textbf{BRIGHT}~\citep{brancati21} consists of breast \glspl*{wsi} from \emph{non-cancerous}, \emph{precancerous}, and \emph{cancerous} subtypes. The slides were acquired at the Fondazione G. Pascale, Italy, and scanned by Aperio AT2 scanner at 40$\times$.
We used the BRIGHT challenge splits\footnote{\url{www.research.ibm.com/haifa/Workshops/BRIGHT}} containing 423, 80, and 200 \glspl*{wsi} for train, validation, and testing.

\textbf{CAMELYON16}~\citep{camelyon16} includes 270 \glspl*{wsi}, 160 normal and 110 with metastases, for training, and 129 slides for testing. The slides were scanned by 3DHISTECH and Hamamatsu scanners at 40$\times$ at the Radboud University Medical Center and the University Medical Center Utrecht, Netherlands.
We split the 270 slides into 90\%/10\% stratified sets for training and validation.

The average number of (pixels, patches), within the tissue area, at 20$\times$ magnification for CRC, BRIGHT, and CAMELYON16 datasets are (227.28 Mpx, 3468), (1.04 Gpx, 15872), and (648.28 Mpx, 9892), respectively.

\subsection{Implementation details}
\label{implementation_details}

\textbf{Preprocessing}:
For each \gls*{wsi}, we detect the tissue area using a Gaussian tissue detector~\citep{jaume21b}, and divide the tissue into patches of size 256$\times$256 at all considered magnifications. We ensure that each high magnification patch is associated to the corresponding lower magnification patch.
We encode the patches with a ResNet-50~\citep{he2016deep} pre-trained on ImageNet~\citep{deng2009imagenet} and apply adaptive average pooling after the third residual block to obtain 1024-dimensional embeddings.

\textbf{$\ours$}:
The gated-attention module comprises three 2-layer \glspl*{mlp}, where the first two are followed by Hyperbolic Tangent and Sigmoid activations, respectively. The classifier is a 2-layer \gls*{mlp} with ReLU activation. We use a dropout probability of 0.25 in all fully-connected layers. 

\textbf{Implementation}:
All methods are implemented using PyTorch~\citep{paszke19} and experiments are run on a single NVIDIA A100 GPU.
$\ours$ uses $K=\{16, 12, 300\}$ on CRC, BRIGHT, and CAMELYON16, respectively, and $\ours$-\textsc{Eff} uses $K=\{12, 8\}$ on CRC and BRIGHT, respectively. We use the Adam optimizer~\citep{kingma15} with $0.0001$ learning rate and plateau scheduler (patience=5 epochs, decay rate=0.8). The experiments are run for 100 epochs with a batch size of one.
The models with the best weighted F1-score (for BRIGHT) and best loss (for CRC \& CAMELYON16) on the validation set are saved for testing.

\subsection{Results and Discussion}\label{sec:results_discussion}
\textbf{Baselines}:
We compare $\ours$ with state-of-the-art \gls*{mil} methods.
Specifically, we compare with $\abmil$~\citep{ilse18}, which uses a gated-attention pooling, and its variant $\clam$~\citep{lu20b}, which further includes an instance-level clustering loss. 
We further compare with two spatially-aware methods, namely, $\transmil$~\citep{shao21} which models instance-level dependencies by pooling based on transformer blocks, and $\sparseconvmil$~\citep{lerousseau2021sparseconvmil} which uses sparse convolutions for pooling. 
For completeness, we also include vanilla \gls*{mil} methods using max-pooling ($\maxmil$)~\citep{lerousseau2021sparseconvmil} and mean-pooling ($\meanmil$)~\citep{lerousseau2021sparseconvmil}.
Notably, we do not compare against multi-scale \gls*{mil} methods, \eg \citep{hashimoto20,li21}, as they encode all patches in a \gls*{wsi} across all considered magnifications.
This is computationally very inefficient, thereby limiting their applicability in practice.
Individual implementation details and hyper-parameters of these baselines are provided in the supplemental material. 
For a fair comparison, preprocessing including the extraction of patch embeddings is done consistently in the same manner, as described in Section~\ref{implementation_details}.

\subsubsection{WSI classification performance}

We present the classification results, in terms of weighted F1-score and accuracy, on CRC, BRIGHT, and CAMELYON16 in Table~\ref{tab:crc}, \ref{tab:bright}, and \ref{tab:camelyon16}. 
Mean$\pm$standard deviation of the metrics are computed over three runs with different weight initializations.
The corresponding magnifications of operation are shown alongside each method for each dataset.
We include two versions of $\ours$ using either 2 or 3 magnifications on CRC and BRIGHT, denoted as $\ours$-\textsc{Eff} (efficient) and $\ours$. 

\newcommand{\std}[1]{{\scriptsize$\pm$#1}}
\begingroup
\setlength{\tabcolsep}{2pt}
\begin{table}[!t]
    \caption{Performance and efficiency measurement on CRC dataset~\citep{oliveira21}. The best and the second best classification results are in \textbf{bold} and \underline{underline}, respectively}
    \centering
    \begin{tabular}{lcccccc}
        \toprule
        \multirow{2}{*}{Methods} &
        \multicolumn{2}{c}{Classification} &
        \multicolumn{2}{c}{Computation} \\
        & Weighted-F1(\%) & Accuracy(\%) & T\glspl*{flop} & Time(s) \\
        \midrule
        $\maxmil$~\citep{lerousseau2021sparseconvmil} (20$\times$) & 82.2\std{0.9} & 82.2\std{1.2} &  0.96 & 0.13 \\
        $\meanmil$~\citep{lerousseau2021sparseconvmil} (20$\times$) & 84.3\std{0.8} & 84.1\std{1.2} & 0.96 & 0.12 \\
        $\sparseconvmil$~\citep{lerousseau2021sparseconvmil} (20$\times$) & 89.6\std{1.3} & 89.6\std{0.9} & 0.96 & 0.13 \\
        $\abmil$~\citep{ilse18} (20$\times$) & 90.1\std{0.6} & 90.2\std{0.5} & 13.63 & 4.85 \\
        $\clam$-SB~\citep{lu20b} (20$\times$) & \underline{90.9\std{0.6}} & \underline{90.9\std{0.5}} & 13.63 & 4.85 \\
        $\transmil$~\citep{shao21} (20$\times$) & 89.8\std{1.1} & 90.2\std{0.9} & 13.63 & 4.85 \\
        \midrule
        $\ours$-\textsc{Eff} (5$\times$ $\rightarrow$ 10$\times$) & 90.3\std{1.3} & 90.3\std{1.3} & 1.06 & 0.38 \\
        $\ours$ (5$\times$ $\rightarrow$ 10$\times$ $\rightarrow$ 20$\times$) & \textbf{92.0\std{0.6}} & \textbf{92.1\std{0.7}} & 1.40 & 0.50 \\
        \bottomrule
    \end{tabular}
    \label{tab:crc}
\end{table}
\endgroup

\begingroup
\setlength{\tabcolsep}{2pt}
\begin{table}[!t]
    \caption{Performance and efficiency measurement on BRIGHT dataset~\citep{brancati21}. The best and the second best classification results are in \textbf{bold} and \underline{underline}, respectively}
    \centering
    \begin{tabular}{lcccccc}
        \toprule
        \multirow{2}{*}{Methods} &
        \multicolumn{2}{c}{Classification} &
        \multicolumn{2}{c}{Computation} \\
        & Weighted-F1 & Accuracy & T\glspl*{flop} & Time(s) \\
        \midrule
        $\maxmil$~\citep{lerousseau2021sparseconvmil} (10$\times$) & 46.8\std{3.7} & 51.3\std{1.7} &  0.96 & 0.13 \\
        $\meanmil$~\citep{lerousseau2021sparseconvmil} (10$\times$) & 44.9\std{2.8} & 47.1\std{0.1} & 0.96 & 0.12 \\
        $\sparseconvmil$~\citep{lerousseau2021sparseconvmil} (10$\times$) & 53.2\std{3.6} & 55.3\std{3.7} & 0.96 & 0.13 \\
        $\abmil$~\citep{ilse18} (10$\times$) & 63.5\std{2.7} & 65.5\std{1.9} & 16.45 & 5.86 \\
        $\clam$-SB~\citep{lu20b} (10$\times$) & 63.1\std{1.7} & 64.3\std{1.7} & 16.45 & 5.86 \\
        $\transmil$~\citep{shao21} (10$\times$) & 65.5\std{2.8} & 66.0\std{2.7} & 16.46 & 5.86 \\
        \midrule
        $\ours$-\textsc{Eff} (1.25$\times$ $\rightarrow$ 2.5$\times$) & \underline{66.0\std{1.9}} & \underline{66.5\std{1.5}} & 0.40 & 0.14 \\
        $\ours$ (1.25$\times$ $\rightarrow$ 2.5$\times$ $\rightarrow$ 10$\times$) & \textbf{68.3\std{1.1}} & \textbf{69.3\std{1.0}} & 1.29 & 0.46 \\
        \bottomrule
    \end{tabular}
    \label{tab:bright}
\end{table}
\endgroup

\begingroup
\setlength{\tabcolsep}{2pt}
\begin{table}[!t]
    \caption{Performance and efficiency measurement on CAMELYON16 dataset~\citep{camelyon16}. The best and the second best classification results are in \textbf{bold} and \underline{underline}, respectively}
    \centering
    \begin{tabular}{lcccccc}
        \toprule
        \multirow{2}{*}{Methods} &
        \multicolumn{2}{c}{Classification} &
        \multicolumn{2}{c}{Computation} \\
        & Weighted-F1(\%) & Accuracy(\%) & T\glspl*{flop} & Time(s) \\
        \midrule
        $\maxmil$\citep{lerousseau2021sparseconvmil} (20$\times$) & 64.0\std{3.0} & 67.1\std{0.9} &  0.96 & 0.13 \\
        $\meanmil$~\citep{lerousseau2021sparseconvmil} (20$\times$) & 63.5\std{1.1} & 65.9\std{1.6} & 0.96 & 0.12 \\
        $\sparseconvmil$~\citep{lerousseau2021sparseconvmil} (20$\times$) & 67.7\std{0.6} & 68.7\std{0.1} & 0.96 & 0.13 \\
        $\abmil$~\citep{ilse18} (20$\times$) & 83.2\std{1.7} & 84.0\std{1.3} & 39.12 & 13.92 \\
        $\clam$-SB~\citep{lu20b} (20$\times$) & 83.3\std{1.5} & 84.0\std{1.3} & 39.12 & 13.92 \\
        $\transmil$~\citep{shao21} (20$\times$) & \textbf{83.6\std{2.6}} & \textbf{85.3\std{1.9}} & 39.12 & 13.92 \\
        \midrule
        $\ours$ (10$\times$ $\rightarrow$ 20$\times$) & \underline{83.3\std{0.3}} & \underline{84.2\std{0.4}} & 14.94 & 5.32 \\
        \bottomrule
    \end{tabular}
    \label{tab:camelyon16}
\end{table}
\endgroup

On CRC, $\ours$ outperforms $\clam$-SB and $\transmil$ by 1.1\% and 2.2\% weighted F1-score, respectively.
Our efficient version achieves performance comparable to $\clam$-SB and $\transmil$, and obtains 1.7\% lower weighted F1 than $\ours$.
For the individual classes, \ie non-neoplastic, low-grade, and high-grade, $\ours$ achieves 94.3\%, 93.6\%, and 86.4\% average F1-scores in the one-vs-rest setting, respectively.

BRIGHT \glspl*{wsi} are 4.5$\times$ larger than CRC \glspl*{wsi}, thus providing better evaluation ground for efficient scaling.
$\ours$ achieves the best performance, outperforming $\clam$-SB and $\transmil$ by 5.2\% and 2.8\% in weighted F1-score. Notably, $\ours$-\textsc{Eff} achieves the second best results by a margin of 2.9\% and 0.5\% over $\clam$-SB and $\transmil$.
For the individual classes, $\ours$ reaches an average F1-score of 70.4\%, 56.5\%, and 77.8\%. The performance is lowest for the pre-cancerous class, which is typically very challenging to identify and often ambiguous with the other two categories.

We also benchmark $\ours$ on the popular CAMELYON16 dataset.
As the metastatic regions can be extremely small (see Figure~\ref{fig:interpretability}), we set the lowest magnification to 10$\times$ in ours.
Nevertheless, this still has an adverse impact on the performance, resulting in 1.1\% lower average accuracy than $\transmil$. However, it translates to misclassifying only one to two \glspl*{wsi} in the test set.

Overall, our proposed method achieves better classification performance on CRC and BRIGHT, while being comparable to the state of the art on CAMELYON16. $\ours$ also consistently outperforms $\ours$-\textsc{Eff}, highlighting the apparent performance-efficiency trade-off, \ie performance reduction in exchange for gains in computational efficiency.

\subsubsection{Efficiency measurements}

We conduct an efficiency analysis, in terms of \glspl*{flop} and processing time for running inference, on CRC, BRIGHT, and CAMELYON16 (see Table~\ref{tab:crc}, \ref{tab:bright}, and \ref{tab:camelyon16}). 
On CRC and BRIGHT, $\ours$ uses $\approx$10$\times$ less \glspl*{flop} and processing time than the best performing baselines, CLAM-SB and TransMIL.
On BRIGHT, our efficient variant reduces computational requirements and time by more than 40$\times$ compared to CLAM-SB and TransMIL while providing comparable performance.
On CAMELYON16, $\ours$ uses approximately a third of the number of \glspl*{flop} required by CLAM-SB and TransMIL. The relatively lower efficiency gain is due to the task itself, where metastases regions include only a small fraction of the \gls*{wsi}, and thus, need to be analysed at a finer magnification. 
On all considered datasets, the methods adopting random patch selection ($\maxmil$, $\meanmil$ and $\sparseconvmil$) highlight similar computational requirements as $\ours$, but perform significantly worse, with >10\% accuracy drop on CAMELYON16 and BRIGHT. 

\begin{figure}[!t]
    \centering
    \includegraphics[width=0.9\textwidth]{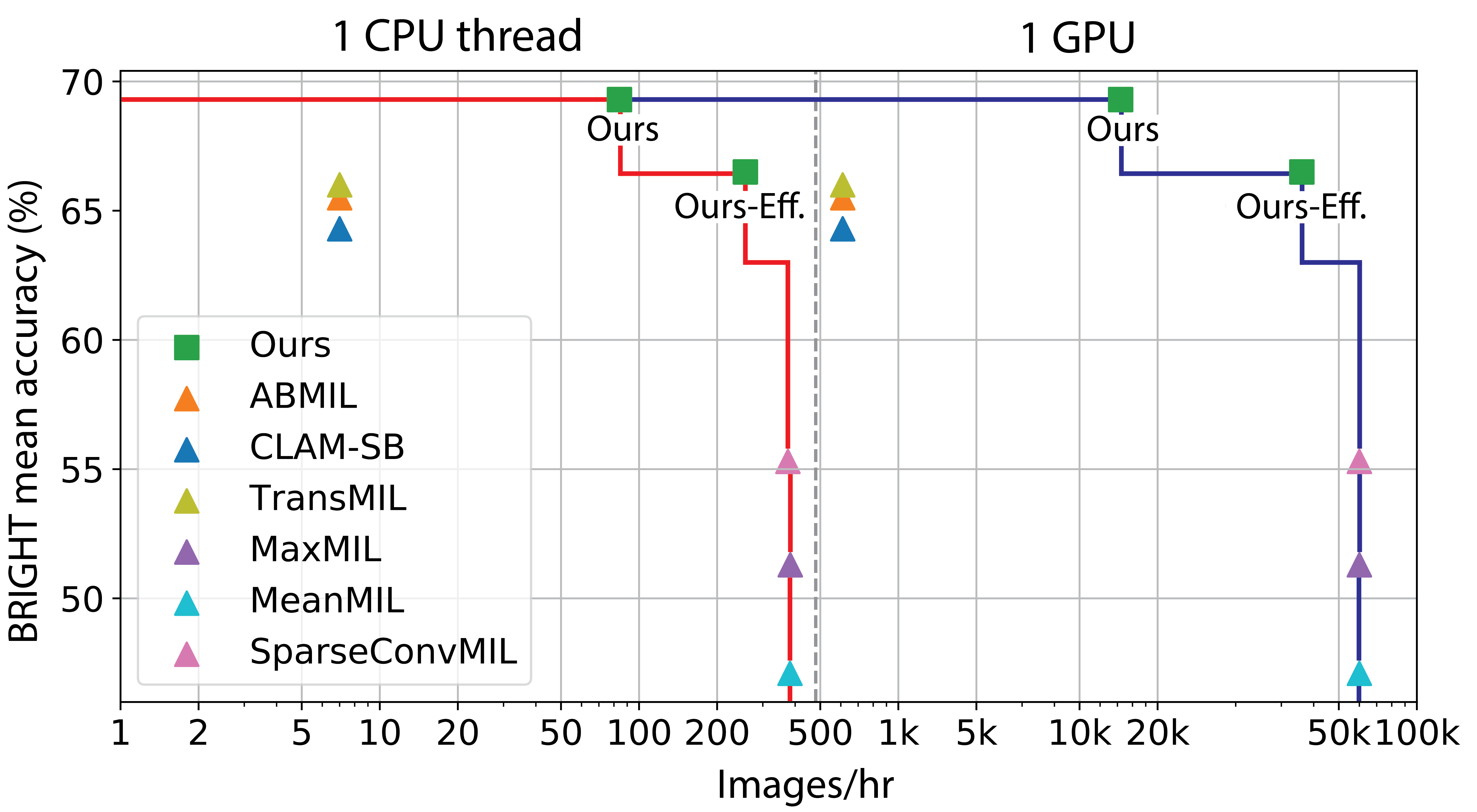}
    \caption{Throughput vs classification accuracy for different \gls*{mil} methods on BRIGHT, (left) on 1 single-core CPU, (right) on 1 NVIDIA A-100 GPU. Efficiency frontier curves are in red and blue for CPU and GPU, respectively}
    \label{fig:time}
    \vspace{-1em}
\end{figure}

To further showcase the efficiency gain of $\ours$, Figure~\ref{fig:time} presents the model throughput (images/hour) against performance (accuracy) for all methods on BRIGHT. The marked efficiency frontier curves signify the best possible accuracies for different minimal throughput requirements.
Noticeably, $\ours$-\textsc{Eff} running on a single-core CPU processor (300 images/h) provides similar throughput to TransMIL and CLAM-SB running on a cutting-edge NVIDIA A100 GPU (600 images/h).
The low computational requirements of $\ours$ makes it more practical and adapted for deployment in clinics, where IT infrastructures are often poorly developed and need large investments for acquiring and maintaining a digital workflow.

\subsubsection{Interpretability}
\label{interpretability}
\begin{figure}[!t]
    \centering
    \includegraphics[width=\textwidth]{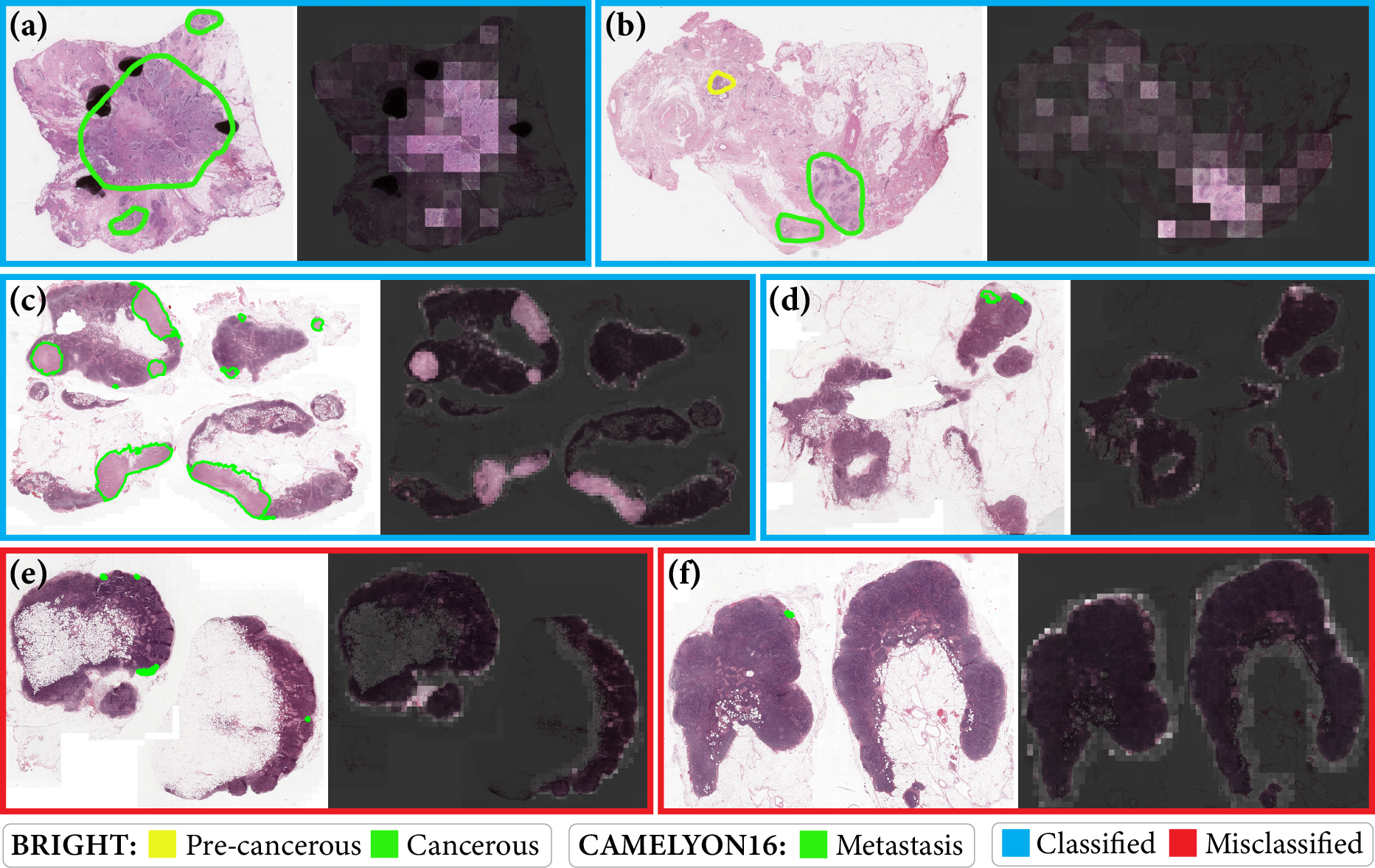}
    \caption{Annotated tumor regions and attention maps from the lowest magnification of $\ours$ are presented for (a,b) BRIGHT and (c-f) CAMELYON16 \glspl*{wsi}} 
    \label{fig:interpretability}
\end{figure}

We interpret $\ours$ by qualitatively analyzing its patch-level attention maps.
Figure~\ref{fig:interpretability}(a,b) demonstrate the maps for two cancerous \glspl*{wsi} in BRIGHT, extracted at 1.25$\times$, and Figure~\ref{fig:interpretability}(c-f) show the maps for four metastatic \glspl*{wsi} in CAMELYON16, extracted at 10$\times$. 
We have further included corresponding tumor regions for each \gls*{wsi}, annotated by an expert pathologist, for comparison. Brighter regions in the attention maps indicate higher attention scores, \ie more influential regions towards the model's prediction.

For the BRIGHT \glspl*{wsi}, $\ours$ correctly attends to cancerous areas in (a,b), pays lower attention to pre-cancerous area in (b), and the least attention to the remaining non-cancerous areas, that include non-cancerous epithelium, stroma, and adipose tissue.
For the CAMELYON16 \glspl*{wsi}, (c,d) are correctly classified as $\ours$ gives high attention to the metastatic regions of different sizes. 
However, the extremely small metastases in (e,f) get low attention, and get disregarded by the Top-K module leading to misclassifying the \glspl*{wsi}.
Notably, for cases with tiny metastases, relatively higher attention is imparted on the periphery of the tissues. This is consistent with the fact that metastases generally appear in the subcapsular zone of lymph nodes, as can be observed in (c-f).
The presented visualizations are obtained from low magnifications in $\ours$, which signifies the learnability of the method to zoom in.
More interpretability maps for other classes, and fine-grained attention maps from higher attention modules in $\ours$ are provided in the supplemental material.

\subsubsection{Ablation study}
We ablated different modules in $\ours$-\textsc{Eff}, due to its simple 2-magnification model. The results on BRIGHT are given in Table~\ref{tab:ablation}.

\begingroup
\setlength{\tabcolsep}{2pt}
\begin{table}[!t]
    \caption{Ablation study on BRIGHT dataset, with the varied algorithmic aspect tabulated in the left-most column. All experiments use $K=8$}
    \centering
    \begin{tabular}{l|lcc}
        \toprule
        & Methods &  Weighted F1(\%) & Accuracy(\%) \\
        \midrule
        \parbox[t]{4mm}{\multirow{4}{*}[-0.05ex]{\rotatebox[origin=c]{90}{Patch sel.}}}
        & $\randomps$ K @ 1.25$\times$ & 61.0 & 61.0 \\
        & $\randomps$ 4K @ 2.5$\times$ & 59.6 & 60.0 \\
        & $\nondiffkps$ K @ 1.25$\times$ & 59.9 & 60.0 \\
        & $\difftopkps$ K @ 1.25$\times$ (Ours) & \textbf{68.1} & \textbf{68.0} \\
        \midrule
        \parbox[t]{4mm}{\multirow{2}{*}[-0.05ex]{\rotatebox[origin=c]{90}{Attn.}}}
        & Single GA @ 1.25$\times$  & 59.6 & 61.0 \\
        & \gls*{dga} @ 1.25$\times$ (Ours)  & \textbf{68.1} & \textbf{68.0} \\
        \midrule
        \parbox[t]{4mm}{\multirow{3}{*}[-0.05ex]{\rotatebox[origin=c]{90}{Feat.}}}
        & Features @2.5$\times$  & 62.7 & 63.5 \\
        & Features @1.25$\times$ $||$ @2.5$\times$  & 64.9 & 65.0 \\
        & Features @1.25$\times$ + @2.5$\times$ (Ours)  & \textbf{68.1} & \textbf{68.0} \\
        \bottomrule
    \end{tabular}
    \label{tab:ablation}
\end{table}
\endgroup

\textbf{Differentiable patch selection}:
We compared our attention-based differentiable patch selection ($\difftopkps$) against three alternatives: random selection at the lowest magnification ($\randomps$ K @ 1.25$\times$), random selection at the highest magnification ($\randomps$ 4K @ 2.5$\times$), and the non-differentiable Top-K selection ($\nondiffkps$) at the lowest magnification.
The top rows in Table~\ref{tab:ablation} show the superiority of $\difftopkps$. It is due to the differentiability of $\difftopkps$, that learns to select patches via the gradient-optimization of the model's prediction.

\textbf{Dual gated attention}:
We examined \gls*{dga} consisting of two separate gated attention modules $\textrm{GA}_{1}$ and $\textrm{GA}'_{1}$ at low magnification, as discussed in Section~\ref{sec:dual_ga}.
The former computes a slide-level representation and the latter learns to select patches at higher magnification.
We can conclude from Table~\ref{tab:ablation} that two separate attentions lead to better patch selection and improved slide representation for overall improved classification.

\textbf{Feature aggregation:}
As shown in Eq.~\eqref{eq:sum_pooling}, we aggregate slide-level representations across magnifications through sum-pooling. Among several alternatives, we compared with: 
using the highest magnification features (Features@2.5$\times$), and
fusing representations via concatenation (herein represented as @1.25$\times$ $||$ @2.5$\times$).
Concatenation improved performance, as shown in Table~\ref{tab:ablation}, indicating the value of multi-scale information.
However, a sum is more natural comparatively, when the two inputs are closely related. 
Our sum-pooling operation is inspired by residual learning, which utilizes the complementarity of two inputs~\citep{he2016deep}. The gain is substantiated in Table~\ref{tab:ablation}, where sum-pooling significantly outperforms concatenation.

\section{Conclusion}
In this work, we introduced $\ours$, a novel framework for \gls*{wsi} classification.
The method is more than an order of magnitude faster than previous state-of-the-art methods during inference, while achieving comparable or better accuracy.
Essential for our method is the concept of differentiable zooming that allows the model to learn which patches are informative and thus worth zooming-in to.
We conduct extensive quantitative and qualitative evaluations on three different datasets, and demonstrate the importance of each component in our model with a detailed ablation study.
Finally, we show that $\ours$ is a modular architecture, that can easily be deployed in different flavors, depending on the performance-efficiency requirements in a given application.

\bibliographystyle{unsrtnat}
\bibliography{references}  

\newpage
\appendix
\section*{Supplemental Material}
\addcontentsline{toc}{section}{Supplemental Material}
\renewcommand{\thesubsection}{\Alph{subsection}}

\noindent In this supplemental material, we include additional results, visualizations, and analyses. The contents of the individual sections are:
\begin{itemize}
    \item Appendix~\ref{app:results}: Confusion matrices for $\ours$ on all datasets. Additional results for the competing methods operating at different magnifications
    \item Appendix~\ref{app:limitations}: Analyzing the limitations of $\ours$
    \item Appendix~\ref{app:impact_of_k}: Analyzing the impact of the number of selected patches in the differentiable Top-K module
    \item Appendix~\ref{app:interpretability}: $\ours$ attention maps at different magnifications
    \item Appendix~\ref{app:implementation}: Training details of the competing baselines 
    \item Appendix~\ref{app:derivation}: Derivation of the Jacobian for differentiable patch selection
\end{itemize}

\subsection{Additional Classification Analysis} \label{app:results}
\begin{figure}[h]
    \centering
    \includegraphics[width=\textwidth]{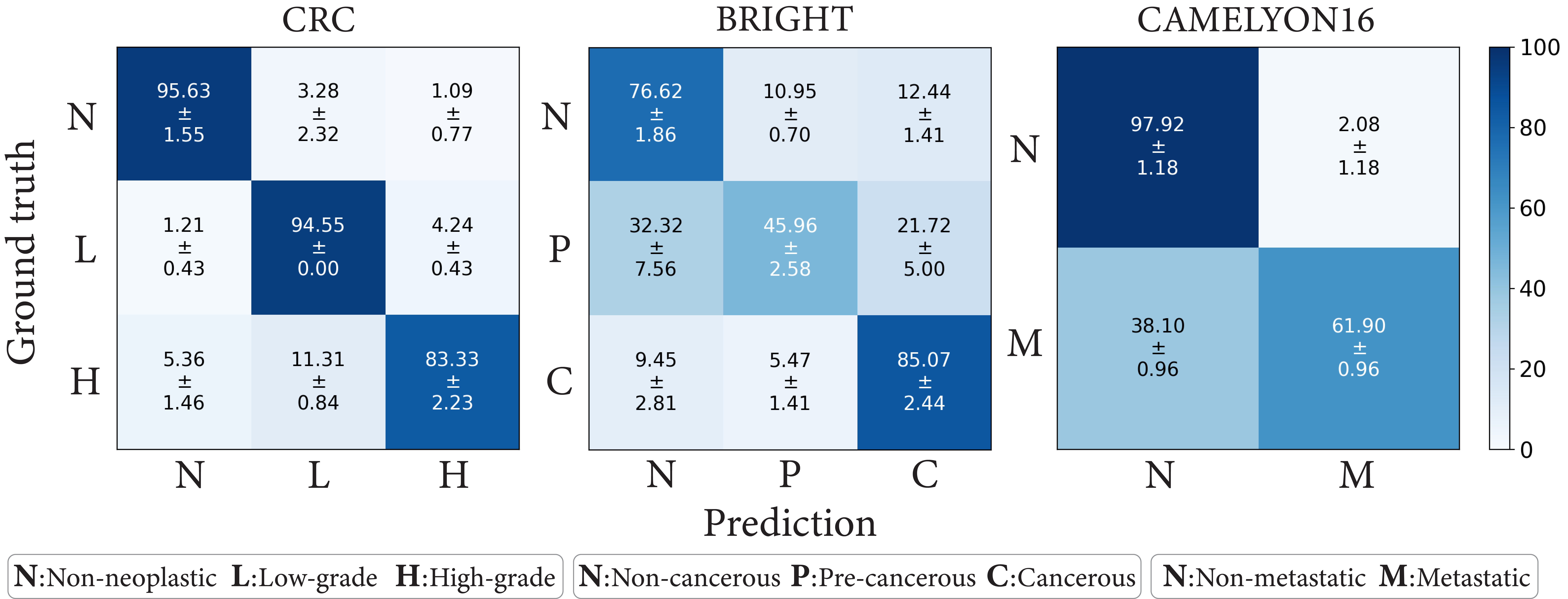}
    \caption{Confusion matrices of $\ours$ on CRC, BRIGHT, and CAMELYON16. 
    An entry represents the corresponding fraction (\%) w.r.t. all samples in the same row}
    \label{fig:sup_conf_mat}
\end{figure}

In Figure~\ref{fig:sup_conf_mat}, we present the confusion matrices of $\ours$ on all datasets.
Results are averaged over three runs with different weight initializations.
On CRC, $\ours$ performs very well and correctly classifies 95.63\% and 94.55\% of the non-neoplastic and low-grade cases, respectively.
Out of all high-grade, cases, our model identifies 83.33\% correctly.
We can see that BRIGHT is the most difficult dataset due to its challenging pre-cancerous class, which is often confused with either the non-cancerous or the cancerous class.
On CAMELYON16, $\ours$ accurately identifies 97.92~\% of non-metastatic cases, while correctly classifying 61.90~\% of metastatic cases.

Additionally, we report the classification performance of all baselines operating at different magnifications in Table~\ref{tab:sup_crc}, \ref{tab:sup_bright}, and \ref{tab:sup_camelyon16}.
It can be observed in Table~\ref{tab:sup_crc} and \ref{tab:sup_bright} that lower magnifications severely impact the performance of the baselines, while $\ours$-\textsc{Eff} performs significantly better.
These results indicate the efficacy of our method, and conclude the benefits of zooming and combining information across magnifications.

\begingroup
\setlength{\tabcolsep}{2pt}
\begin{table}
    \caption{Classification performances on the CRC dataset~\citep{oliveira21}}
    \centering
    \begin{tabular}{lcccc}
        \toprule
        \multirow{2}{*}{Methods} &
        \multicolumn{2}{c}{Classification}\\
        & Weighted-F1(\%) & Accuracy(\%) \\
        \midrule
        $\abmil$~\citep{ilse18} (5$\times$) &86.8\std{0.7} & 87.0\std{0.7}  \\
        $\abmil$~\citep{ilse18} (10$\times$) & 88.8\std{0.7} & 89.0\std{0.6}  \\
        $\clam$-SB~\citep{lu20b} (5$\times$) & 87.7\std{0.5} & 87.8\std{0.5} \\
        $\clam$-SB~\citep{lu20b} (10$\times$) & 89.5\std{0.5} & 89.6\std{0.5} \\
        $\transmil$~\citep{shao21} (5$\times$) & 86.2\std{1.1} & 87.4\std{1.1} \\
        $\transmil$~\citep{shao21} (10$\times$) & 88.4\std{1.3} & 89.1\std{1.1} \\
        \midrule
        $\ours$-\textsc{Eff} (5$\times$ $\rightarrow$ 10$\times$) & 90.3\std{1.3} & 90.3\std{1.3} \\
        $\ours$ (5$\times$ $\rightarrow$ 10$\times$ $\rightarrow$ 20$\times$) & 92.0\std{0.6} & 92.1\std{0.7} \\
        \bottomrule
    \end{tabular}
    \label{tab:sup_crc}
\end{table}
\endgroup

\begingroup
\setlength{\tabcolsep}{2pt}
\begin{table}
    \caption{Classification performances on the BRIGHT dataset~\citep{brancati21}}
    \centering
    \begin{tabular}{lcccc}
        \toprule
        \multirow{2}{*}{Methods} &
        \multicolumn{2}{c}{Classification}\\
        & Weighted-F1(\%) & Accuracy(\%) \\
        \midrule
        $\abmil$~\citep{ilse18} (1.25$\times$) & 58.4\std{1.0} & 58.9\std{1.6}  \\
        $\abmil$~\citep{ilse18} (2.5$\times$) & 58.7\std{1.1} & 59.3\std{1.0}  \\
        $\clam$-SB~\citep{lu20b} (1.25$\times$) & 59.9\std{1.3} & 60.3\std{1.2} \\
        $\clam$-SB~\citep{lu20b} (2.5$\times$) & 60.1\std{1.2} & 60.2\std{1.6} \\
        $\transmil$~\citep{shao21} (1.25$\times$) & 46.1\std{3.8} & 47.3\std{2.5} \\
        $\transmil$~\citep{shao21} (2.5$\times$) & 52.0\std{1.3} & 54.5\std{2.7} \\
        \midrule
        $\ours$-\textsc{Eff} (1.25$\times$ $\rightarrow$ 2.5$\times$) & 66.0\std{1.9} & 66.5\std{1.5}\\
        $\ours$ (1.25$\times$ $\rightarrow$ 2.5$\times$ $\rightarrow$ 10$\times$) & 68.3\std{1.1} & 69.3\std{1.0}\\
        \bottomrule
    \end{tabular}
    \label{tab:sup_bright}
\end{table}
\endgroup

\begingroup
\setlength{\tabcolsep}{2pt}
\begin{table}
    \caption{Classification performances on the CAMELYON16 dataset~\citep{camelyon16}}
    \centering
    \begin{tabular}{lcccc}
        \toprule
        \multirow{2}{*}{Methods} &
        \multicolumn{2}{c}{Classification}\\
        & Weighted-F1(\%) & Accuracy(\%) \\
        \midrule
        $\abmil$~\citep{ilse18} (10$\times$) & 76.7\std{0.8} & 78.3\std{0.7}  \\
        $\clam$-SB~\citep{lu20b} (10$\times$) & 77.5\std{0.6} & 79.1\std{0.6} \\
        $\transmil$~\citep{shao21} (10$\times$) & 76.6\std{1.1} & 79.6\std{1.0} \\
        \midrule
        $\ours$ (10$\times$ $\rightarrow$ 20$\times$) & 83.3\std{0.3} & 84.2\std{0.4} \\
        \bottomrule
    \end{tabular}
    \label{tab:sup_camelyon16}
\end{table}
\endgroup

\subsection{Limitations} \label{app:limitations}
We conjecture that our classification performance on CAMELYON16 is limited by the size of the metastatic regions. For the samples including extremely small metastases, it is challenging to optimize the zooming process.
To validate our hypothesis, we sub-categorize the metastatic samples in the training set into ``small'' and ``large'' metastatic groups, via visual inspection, and create new stratified training and validation sets.
In Table~\ref{tab:limitations}, we present the classification performances of $\ours$ on the validation set, individually for the small and large metastatic samples.
The results show that the performance is significantly higher for large metastases, which substantiates our hypothesis.
\begingroup
\setlength{\tabcolsep}{6pt}
\begin{table}
    \caption{Classification performances on the validation set of CAMELYON16~\citep{camelyon16}.  Validation set is further grouped according to the size of the metastatic regions}
    \centering
    \begin{tabular}{lcccc}
        \toprule
        \multirow{2}{*}{Size of metastases} &
        \multicolumn{2}{c}{Classification}\\
        & Weighted-F1(\%) & Accuracy(\%) \\
        \midrule
        large & 96.2\std{1.0} & 96.1\std{1.1}\\
        small & 86.7\std{1.0} & 87.1\std{1.0}\\
        \bottomrule
    \end{tabular}
    \label{tab:limitations}
\end{table}
\endgroup

\subsection{Impact of K in Differentiable Top-K Patch Selection}
\label{app:impact_of_k}
\begin{figure}[!t]
    \centering
    \includegraphics[width=\textwidth]{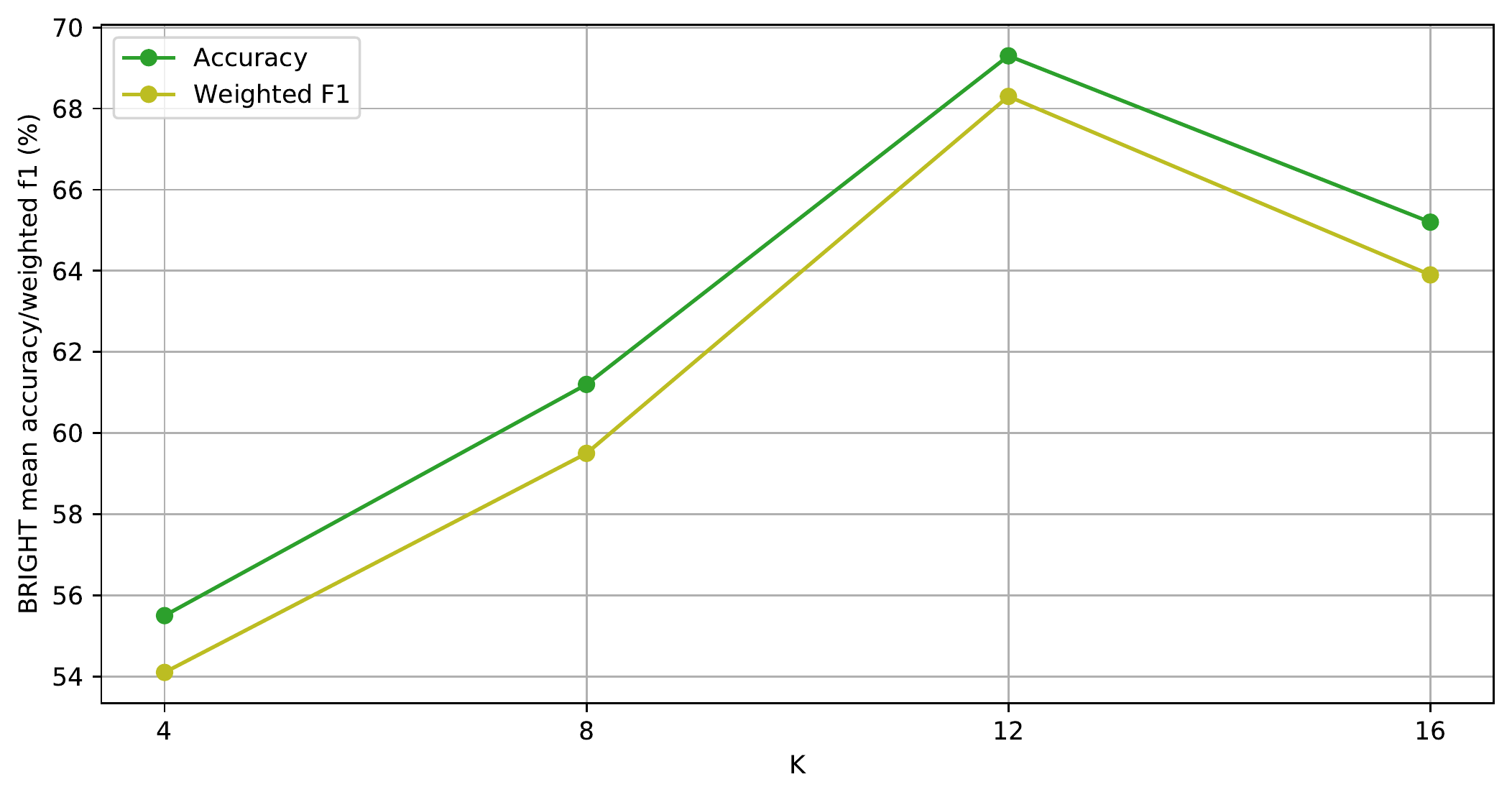}
    \caption{Classification performance of $\ours$ on BRIGHT for different values of $K$}
    \label{fig:diffk}
\end{figure}

Here, we analyze the impact of the number of selected patches ($K$) on the classification performance of $\ours$.
Figure~\ref{fig:diffk} shows that the performance increases with increasing $K$. It peaks at $K=12$ and then slightly drops for further increment.
We reason that this behavior is caused by the average number of patches per \gls*{wsi} being 16 in the BRIGHT dataset at the lowest magnification (1.25$\times$).
Almost all patches are selected in this case, which makes it suboptimal to learn to improve the zooming process.
\begin{figure}[!t]
    \centering
    \includegraphics[width=\textwidth]{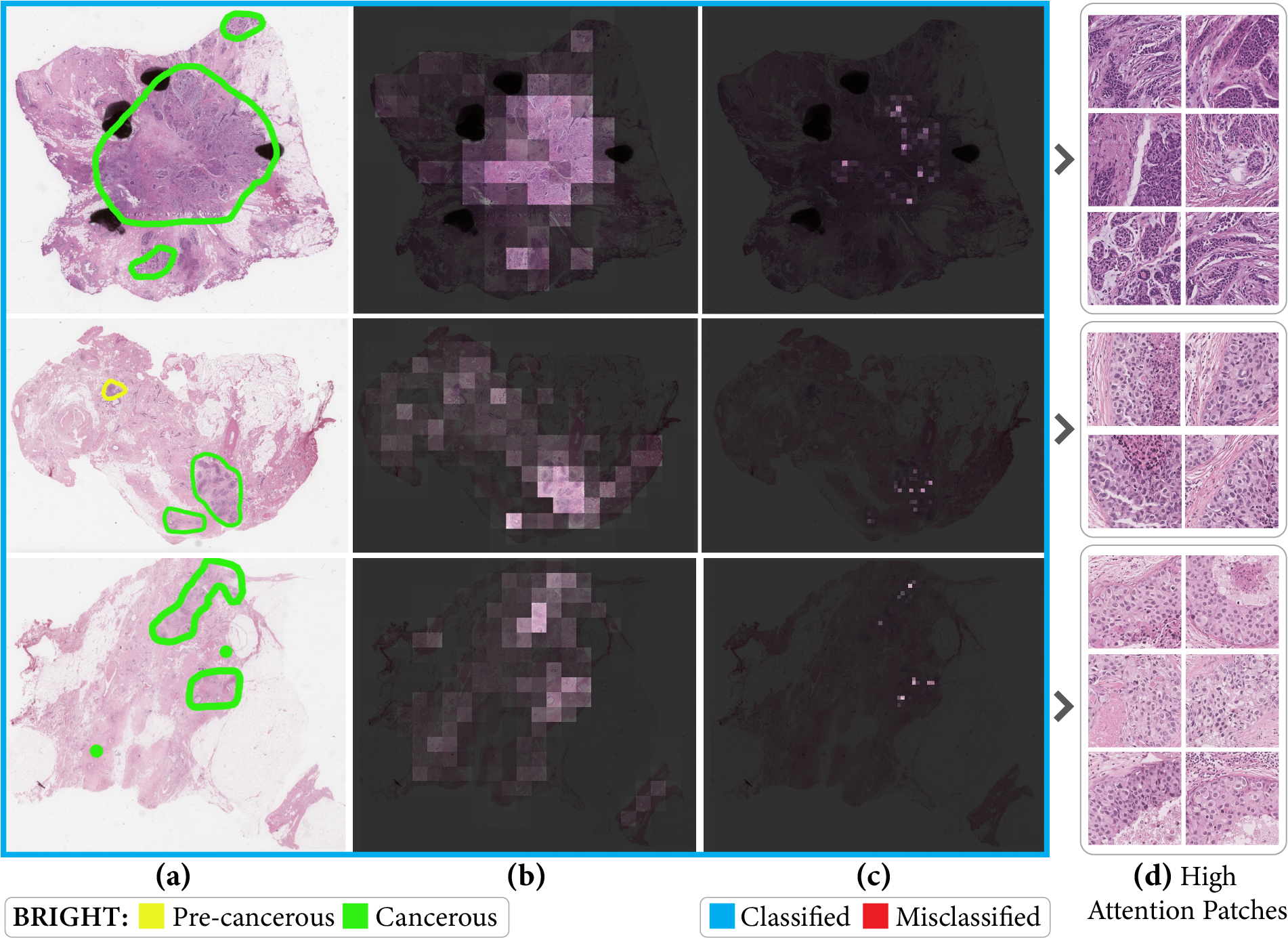}
    \caption{Interpreting $\ours$ on the BRIGHT dataset: (a) annotated tumor regions, (b) attention maps at 1.25$\times$ magnification, (c) attention maps at 10$\times$ magnification, and (d) a subset of extracted patches with high attention scores at 10$\times$ magnification}
    \label{fig:attention}
\end{figure}

\subsection{Interpretability} \label{app:interpretability}
\begin{figure}
    \centering
    \includegraphics[width=0.8\textwidth]{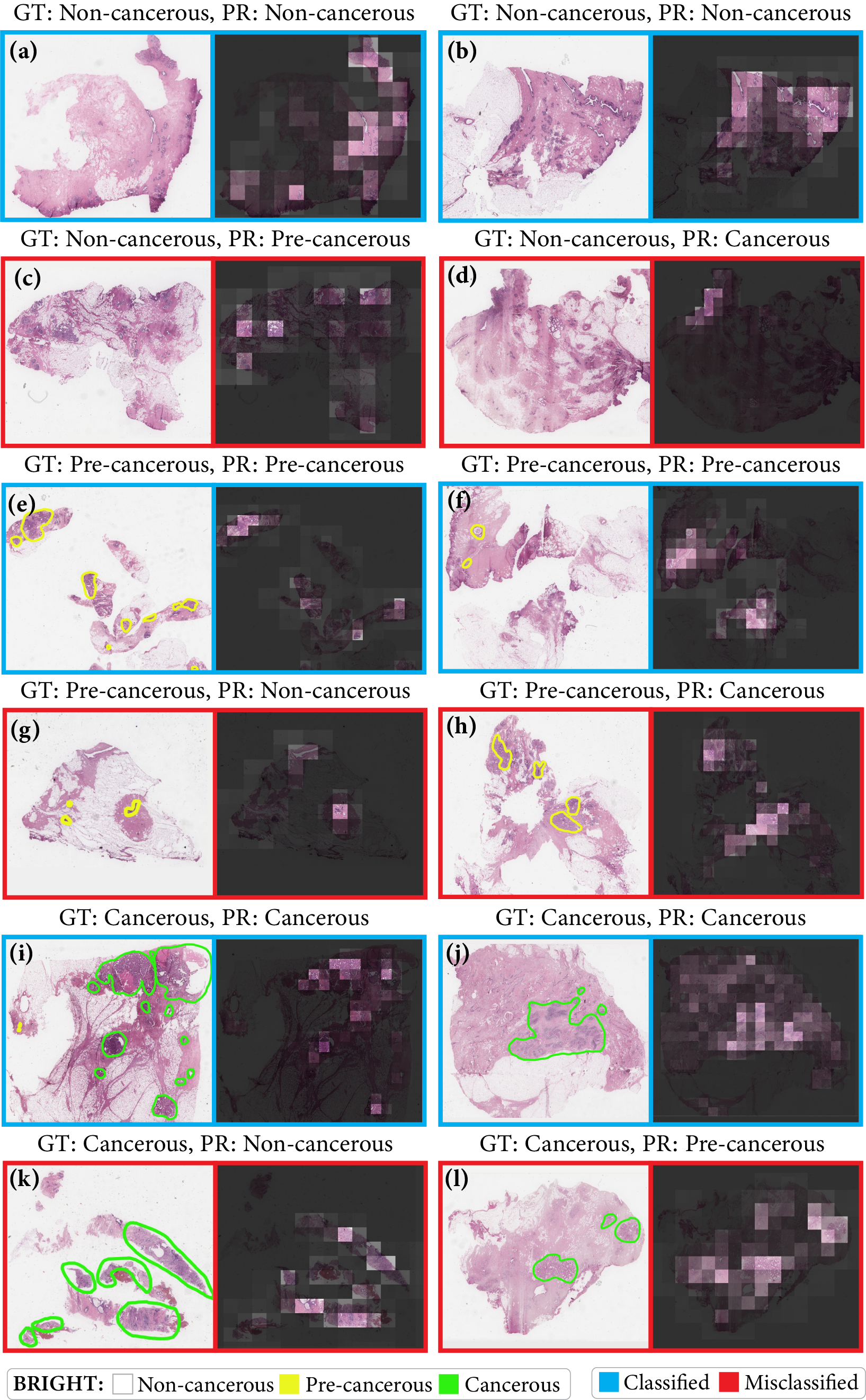}
    \caption{Examples of BRIGHT \glspl*{wsi} with annotated tumor regions and attention maps from $\ours$ at 1.25$\times$ magnification}
    \label{fig:predicted_samples}
\end{figure}

We have shown the $\ours$ attention maps at the lowest magnification, \ie 1.25$\times$, on the BRIGHT dataset in Figure  4 of the main manuscript. 
Here, in Figure~\ref{fig:attention}, we provide the attention maps at both low and high magnification, \ie 1.25$\times$ and 10$\times$, as well as close-up patches with highest attention scores.
We can observe in the zooming process from 1.25$\times$ to 10$\times$, that the model pins down its focus to the most informative regions in the \gls*{wsi}.
The observation is further substantiated by the high attention patches that include cancerous content, \ie invasive tumors and ductal carcinoma in situ tumors, for the presented cancerous \glspl*{wsi}.
Figure~\ref{fig:predicted_samples} shows more examples of annotated tumor regions and attention maps for correctly classified and misclassified samples across all the classes in BRIGHT. For both the correct and incorrect classifications, we can notice that the focus of the model is aligning with the focus of the pathologist. 
However, the morphological ambiguities among the classes finally lead to certain misclassifications.

\subsection{Implementation Details} \label{app:implementation}
\textbf{ABMIL:}
We follow the network architecture proposed in~\citep{ilse18}. The model comprises of a gated attention module consisting of three 2-layer \glspl*{mlp}, where the first two are followed by Hyperbolic Tangent and Sigmoid activations, and a 2-layer \gls*{mlp} classifier with ReLU activation. 

\textbf{CLAM-SB:}
We implement CLAM-SB~\citep{lu20b} with the code\footnote{\url{https://github.com/mahmoodlab/CLAM}} provided by the authors. We use the Adam optimizer~\citep{kingma15} with $0.0001$ learning rate. The maximum and minimum epochs are set to 100, 50, respectively, and use early stopping with patience=$20$ epochs based on the validation weighted F1 score (for BRIGHT) and validation loss (for CRC and CAMELYON16). We use cross-entropy loss for both bag loss and instance loss. The weights of bag-level losses are set to $0.7, 0.5, 0.7$, and the number of positive/negative patches sampled for instance loss are set to $8, 32, 32$ for CRC, BRIGHT and CAMELYON16, respectively. For all three datasets we use a weighted sampler due to class imbalance. 

\textbf{TransMIL:}
We adopt the original implementation\footnote{\url{https://github.com/szc19990412/TransMIL}} of TransMIL~\citep{shao21}. We use Lookahead optimizer~\citep{zhang2019lookahead} with learning rate $0.0002$ and weight decay $0.00001$. The maximum epochs are set to $200$ and early stopping is used with a \emph{patience} of $10$ epochs based on validation weighted F1 score (for BRIGHT) and validation loss (for CRC and Camelyon). We use cross-entropy loss as training loss.

\textbf{SparseConvMIL:}
We adopt the original implementation\footnote{\url{https://github.com/MarvinLer/SparseConvMIL}} of SparseConvMIL~\citep{lerousseau2021sparseconvmil}. 
Unlike other baselines, we run the model on a V100 GPU with 32GB RAM, as it depends on SparseConvNet\footnote{\url{https://github.com/facebookresearch/SparseConvNet}}.
Therefore, we limit the batch size and the number of sampled patches to 8 and 100, respectively.
In the model, we set the number of sparseconv channels to 32, the downsampling factor of the sparse map to 128, and the neurons in the \gls*{mlp} classifier to 128. We use ResNet34 for extracting patch features, and finetune it with learning rate $0.00001$. The learning rate for the rest of the model is $0.001$, and weight decay is $0.0001$. 

\textbf{Max \& Mean MIL:}
We use the formulation presented in \citep{lerousseau2021sparseconvmil} for Max and Mean \textsc{MIL}. We set the same values for the hyperparameters as in SparseConvMIL, but freeze the patch feature extractor, \ie ResNet34.

\subsection{Derivation of Jacobian for Differentiable Patch Selection} \label{app:derivation}
\subsubsection{Perturbed Maximum}
As described in \citep{berthet2020learning}, given a set of distinct points $\mathcal{T} \subset \mathbb{R}^d$ and its convex hull $\mathcal{C}$, a discrete optimization problem with inputs $\mathbf{a}_m \in \mathbb{R}^d$ can generally be formulated as:
\begin{equation}
    \max_{\hat{\mathbf{t}} \in \mathcal{C}} \langle \hat{\mathbf{t}}, \mathbf{a}_m \rangle \qquad \mathbf{t} = \argmax_{\hat{\mathbf{t}} \in \mathcal{C}} \langle \hat{\mathbf{t}}, \mathbf{a}_m \rangle \;.
    \label{eq:opt_problem}
\end{equation}

As per Definition 2.1 in \citep{berthet2020learning}, we can obtain a smoothed $\mathbf{t}$ by adding a random noise vector $\sigma \mathbf{Z} \in \mathbb{R}^d$ with distribution $\mathrm{d}\mu(\mathbf{z}) \propto \exp (-\nu(\mathbf{z}))\mathrm{d}\mathbf{z}$, where $\sigma > 0$ is a scaling parameter.
The perturbed version of the maximizer in Eq.~\eqref{eq:opt_problem} then becomes:
\begin{equation}
    \mathbf{t} = \mathop{\mathbb{E}} \Big[ \argmax_{\hat{\mathbf{t}} \in \mathcal{C}} \langle \hat{\mathbf{t}}, \mathbf{a}_m + \sigma \mathbf{Z} \rangle \Big] \;.
    \label{eq:perturbed_max}
\end{equation}

According to Proposition 3.1 from \citep{berthet2020learning}, the associated Jacobian matrix of $\mathbf{t}$ at $\mathbf{a}_m$ can then be computed as follows:
\begin{equation}
    J_{\mathbf{a}_m} \mathbf{t} = \mathop{\mathbb{E}} \Big[ \argmax_{\hat{\mathbf{t}} \in \mathcal{C}} \langle \hat{\mathbf{t}}, \mathbf{a}_m + \sigma \mathbf{Z} \rangle \nabla_{\mathbf{z}} \nu(\mathbf{Z})^\top / \sigma \Big] \;.
    \label{eq:jacobian_general}
\end{equation}

We choose our noise to have a normal distribution, \ie $\mathbf{Z} \sim \mathcal{N}(0, \mathbbm{1})$. We can thus plug $\nabla_{\mathbf{z}} \nu(\mathbf{Z})^\top = \mathbf{Z}^\top$ into Eq.~\eqref{eq:jacobian_general} and obtain:
\begin{equation}
    J_{\mathbf{a}_m} \mathbf{t} = \mathop{\mathbb{E}}_{\mathbf{Z}\sim \mathcal{N}(0, \mathbbm{1})} \Big[ \argmax_{\hat{\mathbf{t}} \in \mathcal{C}} \langle \hat{\mathbf{t}}, \mathbf{a}_m + \sigma \mathbf{Z} \rangle \mathbf{Z}^\top / \sigma \Big] \;.
    \label{eq:jacobian}
\end{equation}

\subsubsection{Differentiable Top-K Operator}
As shown in~\citep{cordonnier21}, the Top-K selection with sorted indices can be converted into the same form as Eq.~\eqref{eq:opt_problem}.
To this end, the constraint set $\mathcal{C}$ for the indicator matrix $\hat{\mathbf{T}}$ should first be defined as:
\begin{align}
    \mathcal{C} = \Big\{ \hat{\mathbf{T}} \in \mathbb{R}^{N \times K}: \quad \hat{\mathbf{T}}_{n,k} &\geq 0 \\
    \sum_{j=1}^{N} \hat{\mathbf{T}}_{j,k} &= 1 \quad \forall k \in \{1, \dots, K\} \label{eq:col_sum}\\
    \sum_{k=1}^{K} \hat{\mathbf{T}}_{j,k} &\leq 1 \quad \forall j \in \{1, \dots, N\} \label{eq:row_sum}\\
    \sum_{i=1}^{N} i \hat{\mathbf{T}}_{i,k} &< \sum_{j=1}^{N} j \hat{\mathbf{T}}_{j,k'} \quad k < k' \Big\} \label{eq:sort}\;,
\end{align}
where Eq.~\eqref{eq:col_sum} ensures that each column-wise sum in the indicator matrix is equal to one and Eq.~\eqref{eq:row_sum} constrains each row-wise sum to be at most one.
Lastly, Eq.~\eqref{eq:sort} enforces that the indices of the attention weights selected by $\hat{\mathbf{T}}$ are sorted.
With these constraints, the general linear program formulation in Eq.~\eqref{eq:opt_problem} can be used to describe the Top-K selection problem:
\begin{equation}
    \max_{\hat{\mathbf{T}} \in \mathcal{C}} \langle \hat{\mathbf{T}}, \mathbf{a}_m \mathbf{1}^\top \rangle \qquad \mathbf{T} = \argmax_{\hat{\mathbf{T}} \in \mathcal{C}} \langle \hat{\mathbf{T}}, \mathbf{a}_m \mathbf{1}^\top \rangle \;,
    \label{eq:topk_opt_problem}
\end{equation}
where $\mathbf{1}^\top = [1 \cdots 1] \in \mathbb{R}^{1 \times K}$ and thus $\mathbf{a}_m \mathbf{1}^\top \in \mathbb{R}^{N \times K}$ is a matrix containing the attention vectors $\mathbf{a}_m$ repeated $K$ times as its columns. 
Note that here, $\langle \cdot \rangle$ computes the scalar product after vectorizing the matrices.

Now, the perturbed maximizer $\mathbf{T}$ and the corresponding Jacobian $J_{\mathbf{a}_m}\mathbf{T}$ can be computed analogously to Eq.~\eqref{eq:perturbed_max} and Eq.~\eqref{eq:jacobian}, as presented in the main paper.
In contrast to~\citep{berthet2020learning}, however, for the Top-K selection problem, it is required to apply the same noise vector $\sigma \mathbf{Z}$ to each column in $\mathbf{a}_m \mathbf{1}^\top$.
Following the insights from \citep{cordonnier21}, we therefore apply in practice the noise directly to $\mathbf{a}_m$, \ie we employ $\big( \mathbf{a}_m + \sigma \mathbf{Z} \big) \mathbf{1}^\top$ instead of $\mathbf{a}_m \mathbf{1}^\top + \sigma \mathbf{Z}$.

\end{document}